\pdfoutput=1
\documentclass[11pt]{article}

\usepackage[preprint]{acl}

\usepackage{times}
\usepackage{latexsym}

\usepackage[T1]{fontenc}

\usepackage[utf8]{inputenc}

\usepackage{microtype}

\usepackage{inconsolata}

\usepackage{graphicx}

\usepackage[utf8]{inputenc} % allow utf-8 input
\usepackage[T1]{fontenc}    % use 8-bit T1 fonts
\usepackage{hyperref}       % hyperlinks
\usepackage{url}            % simple URL typesetting

\usepackage{amsfonts}       % blackboard math symbols
\usepackage{nicefrac}       % compact symbols for 1/2,  etc.
\usepackage{microtype}      % microtypography
\usepackage{amsmath}
\usepackage{amssymb}
\usepackage{amsthm}
\usepackage[utf8]{inputenc}
\usepackage{graphicx}
\usepackage{colortbl}
\usepackage{bm}
\usepackage{tcolorbox}
\usepackage{multirow} 
\usepackage[ruled,  vlined,  linesnumbered,  commentsnumbered,  longend]{algorithm2e}
\usepackage{listings}
\usepackage{color}
\usepackage{booktabs}
\usepackage{wrapfig} 
\usepackage{graphicx}

\usepackage{subfig}
\usepackage[capitalize]{cleveref}

\usepackage[english]{babel}

\usepackage{amsfonts}
\usepackage{tabularx}
\usepackage{amssymb} 
\usepackage{marvosym}
\usepackage{enumitem}

\usepackage{subcaption}
\usepackage[capitalize]{cleveref}
\usepackage{soul}
\usepackage{xcolor}

\newcommand{\promptfield}[1]{\textcolor{blue}{<#1>}}
\newcommand{\newl}{\textcolor{gray}{\textbackslash n }}

\definecolor{mygreen}{HTML}{38761D}
\definecolor{myred}{HTML}{CC0000}
\definecolor{myblue}{HTML}{56B4E9}
\definecolor{myprefixgreen}{HTML}{009E73}
\definecolor{myorange}{HTML}{D55E00}
\definecolor{mygray}{HTML}{999999}
\definecolor{mydemo}{HTML}{F3C8AD}
\definecolor{myinput}{HTML}{DBE5EE}
\definecolor{mydistill}{HTML}{D6E8CA}
\newcommand{\STAB}[1]{\begin{tabular}{@{}c@{}}#1\end{tabular}}

\newcolumntype{L}{>{\raggedright\arraybackslash}m{4.4cm}}
\newcolumntype{G}{>{\centering\arraybackslash}m{0.5cm}}

\definecolor{customgreen}{HTML}{00B050}

\definecolor{customyellow}{HTML}{BF9000}
\definecolor{customred}{HTML}{FF8B8B}

\usepackage{pifont}% http://ctan.org/pkg/pifont

\theoremstyle{plain}

\theoremstyle{definition}

\theoremstyle{remark}

\definecolor{Gray}{gray}{0.95}
\newcolumntype{a}{>{\columncolor{Gray}}c}

\newcommand{\OURS}{\texttt{AskGNN}}

\title{Let's Ask GNN: Empowering Large Language Model for \\Graph In-Context Learning}

\author{
 \textbf{Zhengyu Hu\textsuperscript{1*}}, 
 \textbf{Yichuan Li\textsuperscript{2*}}, 
 \textbf{ Zhengyu Chen\textsuperscript{3}}, 
 \textbf{Jingang Wang\textsuperscript{3}}, 
\\
 \textbf{ Han Liu\textsuperscript{1}}, 
 \textbf{Kyumin Lee\textsuperscript{2}}, 
 \textbf{Kaize Ding\textsuperscript{1}}
\\
 \textsuperscript{1}Northwestern University, 
 \textsuperscript{2}Worcester Polytechnic Institute, 
 \textsuperscript{3}MeiTuan 
}

\begin{document}
\maketitle
\def\thefootnote{*}\footnotetext{These authors contributed equally. 
}

\begin{abstract}
Textual Attributed Graphs (TAGs) are crucial for modeling complex real-world systems, yet leveraging large language models (LLMs) for TAGs presents unique challenges due to the gap between sequential text processing and graph-structured data. We introduce \OURS{}, a novel approach that bridges this gap by leveraging In-Context Learning (ICL) to integrate graph data and task-specific information into LLMs. \OURS{} employs a Graph Neural Network (GNN)-powered structure-enhanced retriever to select labeled nodes across graphs, incorporating complex graph structures and their supervision signals. Our learning-to-retrieve algorithm optimizes the retriever to select example nodes that maximize LLM performance on graph. Experiments across three tasks and seven LLMs demonstrate \OURS{}'s superior effectiveness in graph task performance, opening new avenues for applying LLMs to graph-structured data without extensive fine-tuning.
\end{abstract}

\section{Introduction}

\begin{figure}[tbh!]
    \centering
    \includegraphics[width=\linewidth]{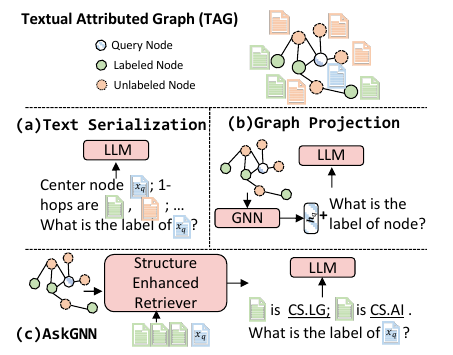}
    \caption{Illustration of methods utilizing LLMs for graph tasks, with a focus on node classification.  Our proposed method enhances structure and task understanding by retrieving insightful examples that improve the LLMs' comprehension of graph complexities.
    }
    \label{fig:illustration}
\end{figure}

Textual attributed graphs (TAGs)~\citep{chen2024exploring, chen2023label,ogb-datasets} are pivotal in modeling complex real-world systems, from social networks to recommendation engines and information retrieval systems. In TAGs, nodes represent text documents, while edges depict relationships between them like shown in \autoref{fig:illustration}. The interconnected nature of TAGs encapsulates rich knowledge and context, offering a more comprehensive understanding of the underlying data structures than isolated text analysis alone. 

Recent advancements in large language models (LLMs)~\citep{brown2020language_gpt, dong2024surveyincontextlearning, hu2024rethinking} have demonstrated remarkable zero-shot and few-shot capabilities across a wide range of tasks. These LLMs excel in areas such as data augmentation~\cite{li2024empowering}, text summarization~\citep{jin2024comprehensivesurveyprocessorientedautomatic}, and content recommendation~\citep{llm-recommendation-survey}. However, despite their impressive capabilities, LLMs usually face significant limitations in processing and leveraging the structural information inherent in TAGs~\citep{yasunaga-etal-2022-linkbert,fatemi2023talk}. 
Trained mainly on unstructured text data, these LLMs lack the innate ability to interpret and utilize the structural information that are fundamental to TAGs. 
This limitation hinders their effectiveness in tasks that require a deeper understanding of interconnections among text documents.

To bridge this gap and empower LLMs for data-efficient graph learning~\cite{zhang2022few,ding2024data}, researchers have developed two primary approaches to integrate structural information into LLM understanding. One line of research, exemplified by studies such as \citet{fatemi2023talk} and \citet{ye2024language}, employs text-based serialization methods to encode $k$-hop neighbors as the context around the query node for LLMs (as illustrated in \autoref{fig:illustration} (a)). However, this method often overlooks nodes beyond the $k$-hop neighborhood, missing critical signals and introducing noise, leading to underperformance compared to Graph Neural Networks (GNNs) in tasks like node classification \citep{chen2024exploring, fatemi2023talk}. 
Another line of research, explored by \citet{chen2024llaga, perozzi2024let, he2024g}, employs GNNs to encode TAGs into compact graph tokens (Fig. \ref{fig:illustration} (b)). While this approach preserves structural information through the message-passing mechanism, it raises the modality misalignment issue between graph structure and text spaces, impeding effective integration with LLMs. These limitations underscore the need for a new solution that can better elicit the power of LLMs on TAGs.

In this paper, we propose {\OURS}, a novel framework that transforms graph structural information and task supervision signals into a limited number of document node-label pairs. 
This approach benefits from LLMs' exceptional power in In-Context Learning (ICL)~\citep{kaplan2020scaling,brown2020language_gpt,li2024mend,dong2024surveyincontextlearning}, leveraging both graph structural information and textual information without extensive fine-tuning. 
At the core of {\OURS} is a GNN-based structure-enhanced retriever, designed to select the most relevant document node-label pairs as ICL examples for LLM few-shot prompting. 
This strategy harnesses GNNs' prowess in extracting structural information while circumventing the semantic gaps between text and non-text tokens \citep{ding2020more, ding2022data, hu2023leveraging}. 
To optimize the retriever, we further introduce a learning-to-retrieve algorithm that fine-tunes the structure-enhanced GNN retriever, ensuring the selection of the most pertinent ICL node examples.  
By integrating these components, {\OURS} offers a new solution that addresses the limitations of previous approaches, enabling LLMs to effectively process and utilize the rich structural and supervisory information inherent in TAGs, while leveraging the adaptability of ICL to diverse tasks with limited supervision, as depicted in \autoref{fig:illustration}(c).
To summarize,  our main contributions are as follows:
\begin{itemize}[leftmargin=*, itemsep=0.5pt, topsep=2.5pt]
    \item We introduce \OURS{}, a new graph in-context learning framework that empowers LLMs for few-shot learning on TAGs.
    \item We develop a learning-to-retrieve algorithm that enhances the retriever's ability to retrieve examples that are both structurally informative and contextually relevant for LLMs' ICL tasks.
    \item We conduct extensive evaluations across three distinct tasks and seven LLMs, demonstrating the superior robustness and effectiveness of \OURS{} in improving graph task performance.
\end{itemize}

\section{Related Work}
\paragraph{LLMs for Graph Learning.}
Applying LLMs to text-attributed graph tasks provides significant opportunities for advancing data-efficient graph learning~\citep{he2023harnessing, li2023grenade, ding2024data, jin2024graph, he2024g}. 
These tasks require LLM to process and understand the complex graph structural information, a challenge that has only recently gained attention~\citep{wang2024instructgraph, ye2024language, huang2024prodigy, he2024g}. 
Some approaches enable LLMs to learn structural information by tuning their parameters~\cite{ye2024language} or through graph instruction tuning~\cite{tang2024graphgptgraphinstructiontuning}. 
Others transform graphs into hidden embeddings to aid in understanding relational data~\citep{he2024g, ye2024language,NEURIPS2023_66178bea}.
However, these methods either cannot fully capture informative graph structural information or align the graph and text tokens.

\paragraph{In-Context Learning.}
ICL~\citep{dong2024surveyincontextlearning, xie2021explanation} allows pretrained LLMs to make predictions for diverse downstream tasks by directly prompting them with a few examples of the task or textual instructions~\citep{Pathak2016ContextEF, min2022rethinking}without finetuning. 
\citet{jiang2023active} improves language models by iteratively retrieving relevant information during text generation,  boosting performance in long-form tasks.
\citet{asai2023self}enhances language models through adaptive retrieval and self-reflection,  outperforming state-of-the-art models in various tasks.
\citet{rubin2021learning, li2023unified, wu2023openicl} train a separate retriever that selects relevant examples using a unified ranking framework,  outperforming task-specific methods.
This capability significantly reduces the adaptation effort compared to traditional fine-tuning approaches and has shown robust performance across a variety of models and tasks \citep{dong2024surveyincontextlearning,  rubin2021learning,  li2024mend}. 
Unlike models such as BERT \citep{devlin2018bert},  which require extensive fine-tuning for new tasks,  LLMs leverage their pretrained knowledge effectively through in-context prompts~\citep{he2024g, ye2024language}.

\section{Problem Definition}
In this work,  we define a TAG  as $\mathcal{G} = (\mathbf{A},  \mathcal{X})$,  where $\mathbf{A} \in \{0,  1\}^{N \times N}$ represents the adjacency matrix,  with $\mathbf{A}_{ij} = 1$ indicating a connection between nodes $i$ and $j$,  $N$ is the total number of nodes and $\mathcal{X}=\{x_i\}_{i=0}^N$ is text corpus for each node. 
Each node $i$ is associated with a text document represented as a sequence of tokens ${x}_i = \{w_v\}_{v=0}^{|x_i|}$. 
Among the nodes in TAG,  only a subset are labeled, denoted as $\mathcal{D} = \{(x_i,  y_i)\}_{i=0}^{M}$,  where $M \ll N$ and $y_i = \{w_v\}_{v=0}^{|y_i|}$ represents the text labels of node $i$ (e.g., 'cs.LG' in the context of arxiv classification).

\begin{figure}[t!]
\centering
\includegraphics[width=0.9\linewidth]{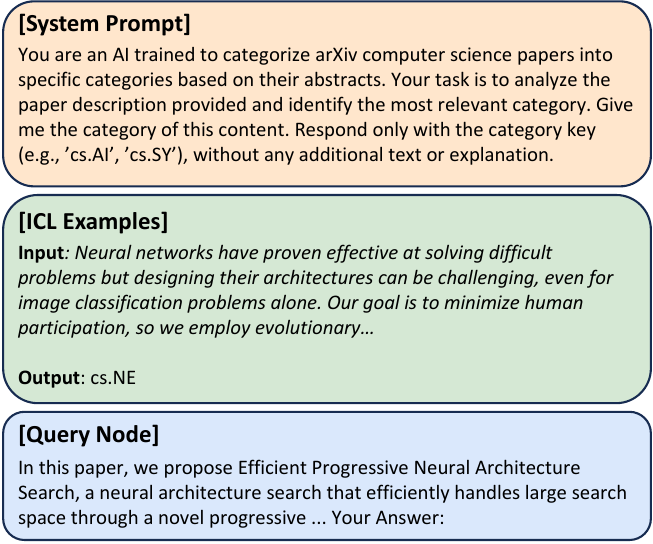}
\caption{Example of node classification task, illustrating how retrieved ICL examples are integrated with the query node with the prompt.}
\label{fig:Promptexample}
\end{figure}

Our objective is to encode both the graph structural information and task supervision signals by retrieving a subset of $K$ labeled examples, $\mathcal{D}_{{q}} = \{(x_i,  y_i)\}_{i \in [K]}$ from $\mathcal{D}$. These examples serve as context for prompting LLMs to address graph-specific tasks for a query node $q$.
This retrieval process,  constrained by both context length and computational costs,  selects far fewer examples than the number of labeled nodes,  with $K \ll N$. The retrieval and prompting process is formalized as follows:
\begin{align}
{\mathcal{D}}_{q} &= R(x_q,  \mathcal{D},  \mathcal{G}), \\
\hat{y}_q &= \text{LLM}(T({\mathcal{D}}_{q},  x_q)), 
\end{align}
where $T$ is the prompt template used to encode both the text from the labeled nodes in ${\mathcal{D}}_{q}$ and the query node $x_q$.  An illustrative example for the node classification task is provided in \autoref{fig:Promptexample}.

\begin{figure*}[tbh!]
    \centering
    \includegraphics[width=1.0\linewidth]{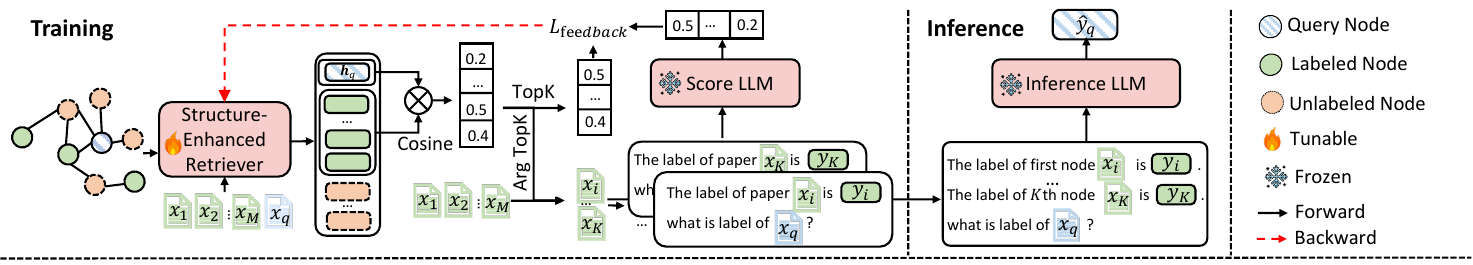}
    \caption{
    The overall framework of \OURS{}, illustrating the structure-enhanced retriever based on GNNs for selecting ICL examples. The framework integrates LLM feedback to optimize the retriever, improving its ability to select relevant examples for graph-based tasks.
    }
    \label{fig:framework}
\end{figure*}

\section{Proposed Approach -- {\OURS}}
\label{sec:Retriever_Enhancement}
We present {\OURS}, a structure-enhanced framework designed to optimize LLMs for graph ICL. The whole framework is illustrated in \autoref{fig:framework}. 
This section details the key components: the GNN-based structure-enhanced retriever (Section \ref{sec:structure_retriever}), LLM feedback collection (Section \ref{sec:icl_rerank}), retriever optimization (Section \ref{sec:optim}), and the utilization of the optimized retriever for ICL example selection (Sectio \ref{sec:R_U}).

\subsection{Structure-Enhanced Retriever}
\label{sec:structure_retriever}

The Structure-Enhanced Retriever (SE-Retriever)
plays a key role in utilizing GNNs to enhance the ICL process for LLMs.  
Previous methods based solely on text similarity \citep{lewis2020retrieval,karpukhin2020dense,xiong2020approximate} fall short when applied to  TAGs, overlooking valuable structural information.
The SE-Retriever is designed to select the most relevant ICL examples from a graph-structured dataset, leveraging both semantic and structural information.
The retrieval process starts with the GNN extracting feature representations from the nodes.  
For a query node $x_q$, its representation at the $l$-th GNN layer is $\mathbf{h}_q^l = \text{GNN}(\mathbf{h}_q^0, \mathbf{A}, \mathbf{H}^0)$, with $\mathbf{h}_q$ denoting the final layer output. The retriever identifies the top $K$ labeled nodes from $M$ labeled nodes based on cosine similarity:
\begin{equation}
\{(x_k, y_k)\}_{k=1}^K =\text{TopK}_{i\in {[M]}} \operatorname{sim}(\mathbf{h}_{q}, \mathbf{h}_{i}).
\end{equation}
The selected ICL examples for the query are formalized as:
\begin{equation}
    \label{eq:ICLSelec}
    \mathcal{D}_{q} =\{(x_i,  y_i)\}_{i \in [K]}. 
\end{equation}
The SE-Retriever ensures that, for each query $x$, the top $K$ samples with the highest cosine similarity are selected, effectively incorporating both textual and structural information into the ICL example selection process.

\subsection{Learning-to-Retrieve via LLM Feedback}
\label{sec:icl_rerank}
We propose a novel Learning-to-Retrieve (L2R) approach that leverages LLM feedback as a training signal to optimize the retriever. This method establishes a dynamic learning loop that continuously refines the retriever's selection of examples based on LLM feedback. The following subsections detail the components and implementation of this learning process.

\paragraph{ICL Training Example Curation.}
Our approach, {\OURS{}} aims to identify which $K$ labeled nodes from the TAGs can serve as effective context for the query node in the LLM. However, evaluating all $K!$ permutations of examples for each query node $x_q$ is computationally prohibitive. To address this, we decompose the problem into evaluating individual examples and collect LLM feedback for each  ICL example separately. To further reduce computational costs and iteratively improve GNN retriever, we only collect feedback for the top-$K$ most similar nodes to each query node, caching the results for efficiency.

\paragraph{LLM Feedback Quantification.}
We introduce a novel "utility score" based on the inverse of perplexity (PPL) to quantify the contribution of a selected example towards the LLM's correct prediction. The utility score for an ICL example $e = (x_e, y_e)$ is defined as:
\begin{equation}
s(e) = \frac{\frac{1}{\text{PPL}(y_q)}}{\sum_{y_c \in \mathcal{Y}} \frac{1}{\text{PPL}(y_c)}}, 
\end{equation}
where $\mathcal{Y}$ represents the set of all possible classes. 
The perplexity for a query $x_q$ with respect to a candidate class $y_c$ and an ICL example $e$ is computed as:
\begin{equation}
\small
\text{PPL}(y_c) = \exp \left\{-\frac{1}{|y_c|} \sum_{v \in [|y_c|]} \log p\left(w_v \mid w_{<v},  x_q,  e \right)\right\}.
\label{eq:ppl}
\end{equation}
This approach allows us to evaluate the contribution of each sample in $\mathcal{D}_{q}$, capturing the LLM's implicit feedback on the effectiveness of different ICL examples.

\paragraph{ICL Example Ranking and Feedback Loss.}
We rank the ICL examples in descending order based on their utility scores:
\begin{equation}
\hat{\mathcal{D}}_{{q}} = \text{Desc.}({ {s(e) \mid e \in \mathcal{D}_q} })
\end{equation}
The resulting ranked set $\hat{\mathcal{D}}_{q}$ serves as a training signal to optimize the GNN retriever, effectively incorporating LLM feedback into the retrieval process. To achieve this optimization, we define  a loss function $\mathcal{L}_{\text{feedback}}$ that focuses on retrieving optimal ICL examples:
\begin{equation}
\mathcal{L}_{\text{feedback}}=- \sum_{q \in [M]}\sum_{k\in [K]} \log  \frac{e^{ \text{sim}\left(\textbf{h}_{q},  \textbf{h}_{{k}}\right)} }{\sum_{j=0}^{K} e^{\text{sim}(\textbf{h}_{q},  \textbf{h}_{{j}}) }}, 
\end{equation}
where ${{k}}$ represents the $k$-th example in $\hat{\mathcal{D}}_{q}$.

\subsection{Optimization}
\label{sec:optim} 
The final loss function combines two components:
\begin{equation}
\label{eq:finalloss}
\mathcal{L} = \beta \times \mathcal{L}_{\text{feedback}}   + (1- \beta) \times \mathcal{L}_{\text{clf}} , 
\end{equation}
where $\mathcal{L}_{\text{feedback}}$ focuses on retrieving optimal ICL examples, and $\mathcal{L}_{\text{clf}}$ targets the graph  learning.
The $\mathcal{L}_{\text{clf}}$ is the node classification loss, defined as:
\begin{equation}
\mathcal{L}_{\text{clf}}=- \sum_{i\in [N]}\text{Cross-Entropy}\left(x_i,  y_{i} \right).
\end{equation}
Optimizing  $\mathcal{L}_{\text{clf}}$ helps the SE-Retriever learn both local and global structural patterns, complementing the $\mathcal{L}_{\text{feedback}}$ to improve example selection and task performance.

\subsection{Model Inference} 
\label{sec:R_U}

\paragraph{ICL Example Selection.} 
The SE-Retriever  $R$ is employed to select an optimal set of ICL examples from the labeled TAGs. 
This selection is driven by the learned structural information and the similarity between the query $x_q$ and the potential examples:
\begin{equation}
\{(x_i,  y_i)\}_{i \in [K]}=  R(x_q,  \mathcal{D},  \mathcal{G}).
\end{equation}

\paragraph{Node Classification.} 
The LLM leverages the selected ICL examples $\hat{\mathcal{D}}_{{q}} = \{(x_i,  y_i)\}_{i \in [K]}$ 
alongside the query $x_q$ to perform the classification task.
This process is formalized as follows:
\begin{equation}
\hat{y}_q = \text{LLM}(T(\hat{\mathcal{D}}_{q},  x_q)), 
\end{equation}
where $T$ is the prompt template used to encode both the labeled nodes in $\hat{\mathcal{D}}_{q}$ and the text of the query node $x_q$.
By integrating the most relevant examples, this approach enhances the LLM’s understanding, resulting in improved prediction accuracy for our tasks.

\section{Experiment}

\begin{table*}[h]
\centering
\caption{
Averaged node classification accuracy on the \texttt{ogbn-arxiv}, \texttt{ogbn-products}, and \texttt{arxiv2023} datasets with 1\%, 3\%, and 10\% of labeled nodes. For each LLM family, the best results are \textbf{bold}.
}
\scalebox{0.67}{
  
\begin{tabular}{c lll lll lll lll c}
\toprule
&  \multirow{2}{*}{\textbf{Methods}} & \multicolumn{3}{c}{\textbf{ogbn-arxiv}}                                                                       & \multicolumn{3}{c}{\textbf{ogbn-products}}       & \multicolumn{3}{c}{\textbf{Arxiv2023}}                                                              & \multirow{2}{*}{\textbf{Avg.}} \\ \cmidrule(l){3-11} 
&  & \multicolumn{1}{c}{1\%}     & \multicolumn{1}{c}{5\%}   & \multicolumn{1}{c}{10\%}  & \multicolumn{1}{c}{1\%}   & \multicolumn{1}{c}{5\%}   & \multicolumn{1}{c}{10\%}  & \multicolumn{1}{c}{1\%}     & \multicolumn{1}{c}{5\%}   & \multicolumn{1}{c}{10\%}   &    \\ \midrule
 \multirow{3}{*}{\STAB{\rotatebox[origin=c]{90}{{N/A}}}} & GCN &  57.56 \scriptsize$\pm$4.18        & 65.91 \scriptsize$\pm$1.65        & {67.77 \scriptsize$\pm$1.75}        & 68.21 \scriptsize$\pm$1.75        & 73.06 \scriptsize$\pm$1.50        & {74.45 \scriptsize$\pm$1.45}        & 53.41 \scriptsize$\pm$3.42        & 61.91 \scriptsize$\pm$1.60        &{64.66 \scriptsize$\pm$1.48}        &   65.22  \\
  & GraphSAGE                                                    & 57.72 \scriptsize$\pm$1.66     & 62.20 \scriptsize$\pm$2.23   & 65.71 \scriptsize$\pm$2.33  & 65.72 \scriptsize$\pm$1.66  & 71.32  \scriptsize$\pm$1.80  & 72.45 \scriptsize$\pm$1.75  & 54.67 \scriptsize$\pm$2.14        & 61.71 \scriptsize$\pm$1.50        & 64.51 \scriptsize$\pm$1.50  & 64.00        \\ 
  & GraphSAINT & 58.03 \scriptsize$\pm$2.17     & 62.73 \scriptsize$\pm$3.03   & 65.92 \scriptsize$\pm$2.78  & 66.13 \scriptsize$\pm$1.53  & 71.73  \scriptsize$\pm$1.23  & 73.54 \scriptsize$\pm$1.18  & 54.92 \scriptsize$\pm$2.07       & 61.95 \scriptsize$\pm$1.79        & 64.84 \scriptsize$\pm$1.94  &    64.42 \\
  \midrule
\multirow{8}{*}{\STAB{\rotatebox[origin=c]{90}{{Qwen1.5}-72B}}}                                             & K-Hop & 62.21 \scriptsize$\pm$0.22  & 62.21 \scriptsize$\pm$0.22 & 62.21 \scriptsize$\pm$0.22& 66.85 \scriptsize$\pm$0.15 & 66.85 \scriptsize$\pm$0.15 & 66.85 \scriptsize$\pm$0.15  & 67.47 \scriptsize$\pm$0.24  & 67.47 \scriptsize$\pm$0.24  & 67.47 \scriptsize$\pm$0.24  &  65.29  \\                                  
  & Graph-CoT & 59.73 \scriptsize$\pm$0.21  & 59.73 \scriptsize$\pm$0.21 & 59.73 \scriptsize$\pm$0.21 & 67.78 \scriptsize$\pm$0.17 & 67.78 \scriptsize$\pm$0.17 & 67.78 \scriptsize$\pm$0.17  & 64.89 \scriptsize$\pm$0.13        & 64.89 \scriptsize$\pm$0.13        & 64.89 \scriptsize$\pm$0.13  & 64.13 \\ 
  \cmidrule{2-12}
  & InstructTuning       & {{65.33}} \scriptsize$\pm$0.35                & {{70.63}} \scriptsize$\pm$0.76        & {{71.47}} \scriptsize$\pm$0.58        & 73.70 \scriptsize$\pm$0.53           & 80.16 \scriptsize$\pm$0.41        & {{81.99}} \scriptsize$\pm$0.26    & {{69.32}} \scriptsize$\pm$0.23  & {{70.72}} \scriptsize$\pm$0.33  & {{71.23}} \scriptsize$\pm$0.25    & {{72.72}}     \\
  & InstructGLM      & {\textbf{65.55}} \scriptsize$\pm$0.26               & {\textbf{70.81}} \scriptsize$\pm$0.62        & {\textbf{71.51}} \scriptsize$\pm$0.36        & {{73.83}} \scriptsize$\pm$0.48           & {\textbf{80.28}} \scriptsize$\pm$0.51        & {\textbf{82.12}} \scriptsize$\pm$0.24    & 69.31 \scriptsize$\pm$0.19  & {\textbf{70.86}} \scriptsize$\pm$0.28  & {\textbf{71.31}} \scriptsize$\pm$0.21  & \textbf{{72.84}}  \\                               
  \cmidrule{2-12}
   & Zero-Shot                                               & 61.63 \scriptsize$\pm$0.30  & 61.63 \scriptsize$\pm$0.30 & 61.63 \scriptsize$\pm$0.30 & 62.85 \scriptsize$\pm$0.27 & 62.85 \scriptsize$\pm$0.27 & 62.85 \scriptsize$\pm$0.27 & 67.76 \scriptsize$\pm$0.53  & 67.76 \scriptsize$\pm$0.53  & 67.76 \scriptsize$\pm$0.53  & 64.08  \\
& Few-Shot (Rand.) & 61.90 \scriptsize$\pm$0.02  & 62.30 \scriptsize$\pm$0.14 & 62.17 \scriptsize$\pm$0.31 & 66.80 \scriptsize$\pm$0.04 & 67.18 \scriptsize$\pm$0.27 & 67.62 \scriptsize$\pm$0.08  & 67.56 \scriptsize$\pm$0.23        & 67.66 \scriptsize$\pm$0.69        & \multicolumn{1}{c}{68.16 \scriptsize$\pm$0.23}  & 65.70  \\
  & Few-Shot ($k$-NN) & 63.47 \scriptsize$\pm$0.37  & 64.77 \scriptsize$\pm$0.12 & 64.87 \scriptsize$\pm$0.35 & 74.58 \scriptsize$\pm$0.39 & 75.21 \scriptsize$\pm$0.21 & 76.48 \scriptsize$\pm$0.22  & 67.47 \scriptsize$\pm$0.57        & 68.06 \scriptsize$\pm$0.09        & \multicolumn{1}{c}{68.86 \scriptsize$\pm$1.52}  &  69.30  \\
\rowcolor[HTML]{EFEFEF}  & {\textbf{{\OURS}}}                                       & 64.67 \scriptsize$\pm$0.10  & 66.95 \scriptsize$\pm$0.23 & 67.82 \scriptsize$\pm$0.16 & {\textbf{76.79}} \scriptsize$\pm$0.26 & {{78.72}} \scriptsize$\pm$0.19 & 79.91 \scriptsize$\pm$0.23   & {\textbf{69.56}} \scriptsize$\pm$0.22        & 69.97 \scriptsize$\pm$0.12        & \multicolumn{1}{c}{70.46 \scriptsize$\pm$0.17}  & \textbf{71.65} \\ \midrule

 \multirow{8}{*}{\STAB{\rotatebox[origin=c]{90}{{Llama3}-70B}}}  & K-Hop & 66.27 \scriptsize$\pm$0.28          & 66.27 \scriptsize$\pm$0.28        & 66.27 \scriptsize$\pm$0.28        & 75.93 \scriptsize$\pm$0.29      & 75.93 \scriptsize$\pm$0.29        & 75.93 \scriptsize$\pm$0.29     & 68.23 \scriptsize$\pm$0.19  & 68.23 \scriptsize$\pm$0.19 & 68.23 \scriptsize$\pm$0.19 &    70.14       \\
  & Graph-CoT                                                   &  64.51 \scriptsize$\pm$0.21         &  64.51 \scriptsize$\pm$0.21       &  64.51 \scriptsize$\pm$0.21        & 77.52 \scriptsize$\pm$0.13        & 77.52 \scriptsize$\pm$0.13        & 77.52 \scriptsize$\pm$0.13       & 64.31 \scriptsize$\pm$0.36  & 64.31 \scriptsize$\pm$0.36  & 64.31 \scriptsize$\pm$0.36   &    68.84         \\
  \cmidrule{2-12}
  & InstructTuning       & 68.30 \scriptsize$\pm$0.10        & {{71.10}} \scriptsize$\pm$0.30        & 71.07 \scriptsize$\pm$0.67        & 72.20 \scriptsize$\pm$0.89              & 81.87 \scriptsize$\pm$0.27        & 82.42 \scriptsize$\pm$0.21    & 68.96 \scriptsize$\pm$0.24  & 69.43 \scriptsize$\pm$0.24  & 69.52 \scriptsize$\pm$0.22  & 72.76 \\
& InstructGLM      & {{68.35}} \scriptsize$\pm$0.16        & {\textbf{71.17}} \scriptsize$\pm$0.25        & {{71.13}} \scriptsize$\pm$0.42        & 72.19 \scriptsize$\pm$0.77              & {{81.70}} \scriptsize$\pm$0.29        & {{82.59}} \scriptsize$\pm$0.32    & {{69.02}} \scriptsize$\pm$0.17  & {{69.57}} \scriptsize$\pm$0.13  & {{69.61}} \scriptsize$\pm$0.32  & {{72.82}}  \\
\cmidrule{2-12}
& Zero-Shot & 68.43 \scriptsize$\pm$0.31              & 68.43 \scriptsize$\pm$0.31        & 68.43 \scriptsize$\pm$0.31        & 77.00 \scriptsize$\pm$0.16         & 77.00 \scriptsize$\pm$0.16        & 77.00 \scriptsize$\pm$0.16     & 69.03 \scriptsize$\pm$0.22  & 69.03 \scriptsize$\pm$0.22  & 69.03 \scriptsize$\pm$0.22     &   71.48         \\
  & Few-Shot (Rand.)                                                & 66.27 \scriptsize$\pm$0.21         & 66.10 \scriptsize$\pm$0.17        & 66.20 \scriptsize$\pm$0.17        & 76.87 \scriptsize$\pm$0.18        & 76.86 \scriptsize$\pm$0.21        & 77.00 \scriptsize$\pm$0.26      & 68.21 \scriptsize$\pm$0.23  & 68.35 \scriptsize$\pm$0.16  & 68.18 \scriptsize$\pm$0.28   &  70.44        \\
  & Few-Shot ($k$-NN)                                       & 68.23 \scriptsize$\pm$0.25         & 67.80 \scriptsize$\pm$0.10        & 67.63 \scriptsize$\pm$0.35        & {{78.76}} \scriptsize$\pm$0.39        & 79.00 \scriptsize$\pm$0.10        & 79.19 \scriptsize$\pm$0.18   & 68.36 \scriptsize$\pm$0.17  & 68.47 \scriptsize$\pm$0.14  & 68.25 \scriptsize$\pm$0.18     & 71.74        \\
\rowcolor[HTML]{EFEFEF} & {\textbf{{\OURS}}}   & {\textbf{69.13}} \scriptsize$\pm$0.24             & 70.29 \scriptsize$\pm$0.25        & {\textbf{71.53}} \scriptsize$\pm$0.11        & {\textbf{81.35}} \scriptsize$\pm$0.23              & {\textbf{82.54}} \scriptsize$\pm$0.13        & {\textbf{82.88}} \scriptsize$\pm$0.17      & {\textbf{73.86}} \scriptsize$\pm$0.22        & {\textbf{74.07}} \scriptsize$\pm$0.12        & \multicolumn{1}{c}{{\textbf{74.97}} \scriptsize$\pm$0.17}  &  \textbf{{73.86}}  \\  
\bottomrule
\end{tabular}
}
\label{tab:nc_result}
\end{table*}

\subsection{Experimental Setup}

\paragraph{Evaluation Datasets.}
In this paper,  we adopt the following TAG datasets widely used for node classification: ogbn-arxiv,  ogbn-products~(OGB)~\citep{ogb-datasets} and arxiv2023~\citep{he2023harnessing}, 
The statistics and descriptions of these datasets are provided in~\autoref{sec:dataset}. 
To assess the effectiveness of {\OURS} under low-data scenarios, we limit the training dataset to 1\%, 5\%, and 10\% of the labeled nodes.
We use Accuracy as the evaluation metric in all experiments and employ the default numerical node embeddings as document node representation.

\paragraph{Baseline Methods.} 
The baseline methods in our study fall into the following categories:
\textit{(1)} Bare GNNs: GCN~\cite{kipf2017semisupervised}, GraphSAGE~\cite{hamilton2017inductive_sage} and GraphSAINT~\cite{zeng2019graphsaint}.
\textit{(2)} Text-based serialization: \textit{K-Hop}~\citep{chen2024exploring},  \textit{Graph-CoT}~\citep{jin2024graph});
\textit{(3)} Graph projection: \textit{InstructGLM}~\citep{ye2024language};
\textit{(4)} ICL: Zero-shot,  randomly sampling based few-shot and semantic-based $k$-NN few-shot;
\textit{(5)} InstructTuning~\citep{zheng2024llamafactory}.
To show the superiority of our approach,  we first include both widely used and state-of-the-art methods as our baselines.
A detailed description of these methods can be found in ~\autoref{sec:baselineDetail}.

\paragraph{Implementation Details.} 
\label{sec:imple-detail}

We adopted Qwen1.5-72B~\citep{bai2023qwen} and Llama3-70B~\citep{llama3modelcard} as our primary inference LLMs in the main experiments.
For scoring, we utilize smaller models like Qwen1.5-7B and Llama-8B to complement the larger inference LLMs.
Additional model architectures, such as Mistral-8x7B~\citep{mistral87bmodelcard}, and varying model scales, including 7B, 14B, and 32B, are tested in Section~\ref{sec:diffSize}. 
We use GraphSAGE~\cite{hamilton2017inductive_sage} as our retriever backbone.

\subsection{Experiment Results}
\paragraph{Main Results.} To evaluate the overall performance of {\OURS}, we conducted experiments across three datasets.
The results presented in \autoref{tab:nc_result} highlight several key findings:
First, our approach consistently outperforms other methods across all datasets and baselines, demonstrating its effectiveness in selecting optimal examples for ICL.
Second, the correct selection of ICL examples significantly enhances LLM performance, whereas poorly chosen ICL examples can have detrimental effects.
For instance, with the Llama3-70B model, Few-Shot learning performs worse than Zero-Shot. However, when the ICL examples are selected using {\OURS}, the model’s performance improves markedly. Similar conclusions are drawn for Few-shot (Rand.) and Few-shot ($k$-NN).
Third, we observe that the impact of Instruct Tuning is correlated with the power of the LLM. 
Specifically, when the LLM is more powerful (e.g., Llama3-70B), the gains from Instruct Tuning are smaller compared to those observed with weaker models (e.g., Qwen1.5-72B). 
This may be because Llama3 has undergone extensive training on relevant tasks~\citep{llama3modelcard}.
Finally, consistent with previous studies~\citep{chen2024exploring}, text-based serialization methods do not show significant improvement in text-attributed graph  tasks compared to Zero-Shot methods. 
In graph datasets, these methods are less effective than ICL methods that effectively leverage structural information.

\begin{figure}[h!]
    \centering
    \includegraphics[width=1.0\columnwidth]{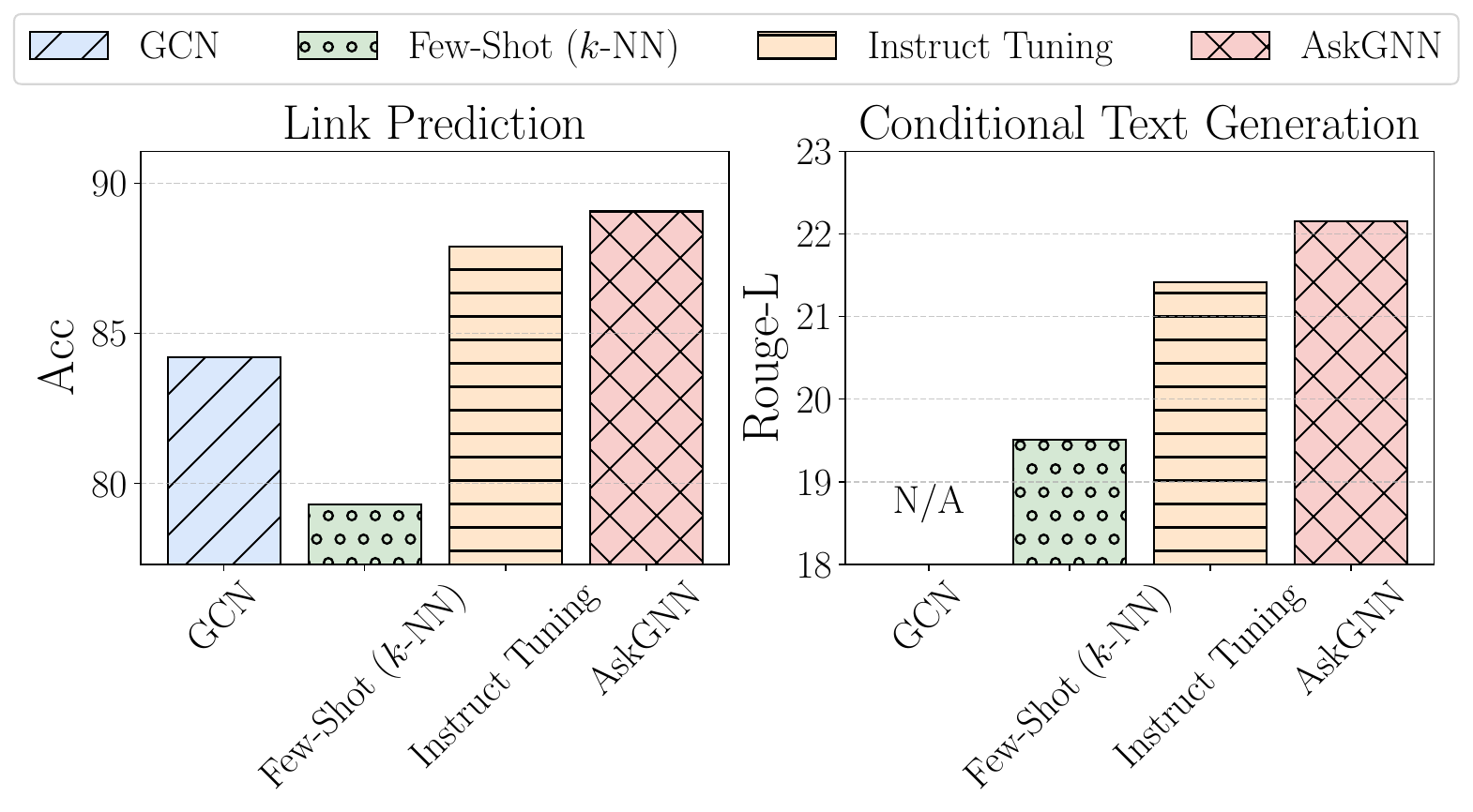}
\caption{
Performance of \OURS{} on different tasks, including link prediction and conditional text generation.
}
\label{fig:MoreTask}
\end{figure}

\paragraph{Extended Results.} We expanded the evaluation of {\OURS{}} to diverse tasks such as link prediction and conditional text generation, demonstrating its broader applicability and versatility in addressing various graph-based challenges.
The results of link prediction and conditional text generation can be found in \autoref{fig:MoreTask}. 
% \paragraph{Link Prediction.}
For the link prediction task, we reformulated the problem as binary question classification.
In this setting, given a pair of nodes, the model predicts whether a connection exists between them. 
In our experiments, \OURS{} outperformed baseline methods, achieving 89.06\% accuracy, surpassing both GCN and instruction-tuned models.
In the conditional text generation task, we evaluated \OURS{}'s ability to integrate graph knowledge into language generation. 
The model was given with 10\% of a query node’s text and tasked with generating the remaining 90\%. 
Using Rouge-L~\citep{lin-2004-rouge} as the evaluation metric, \OURS{} achieved a score of 22.15, surpassing both few-shot learning and instruction-tuned baselines.
These results underscore \OURS{}’s flexibility and adaptability in many other tasks besides node classification.

\section{Further Analysis}

\subsection{Analysis on LLMs' Sizes and Families}
\label{sec:diffSize}
This section aims to provide a comprehensive understanding of how varying inference LLM's sizes and architectures impact the performance of \OURS{} in integrating semantic and structural information from text-attributed graphs. 

\paragraph{Same Family but Different Sizes.}

\begin{figure}
    \centering
    \includegraphics[width=1.0\linewidth]{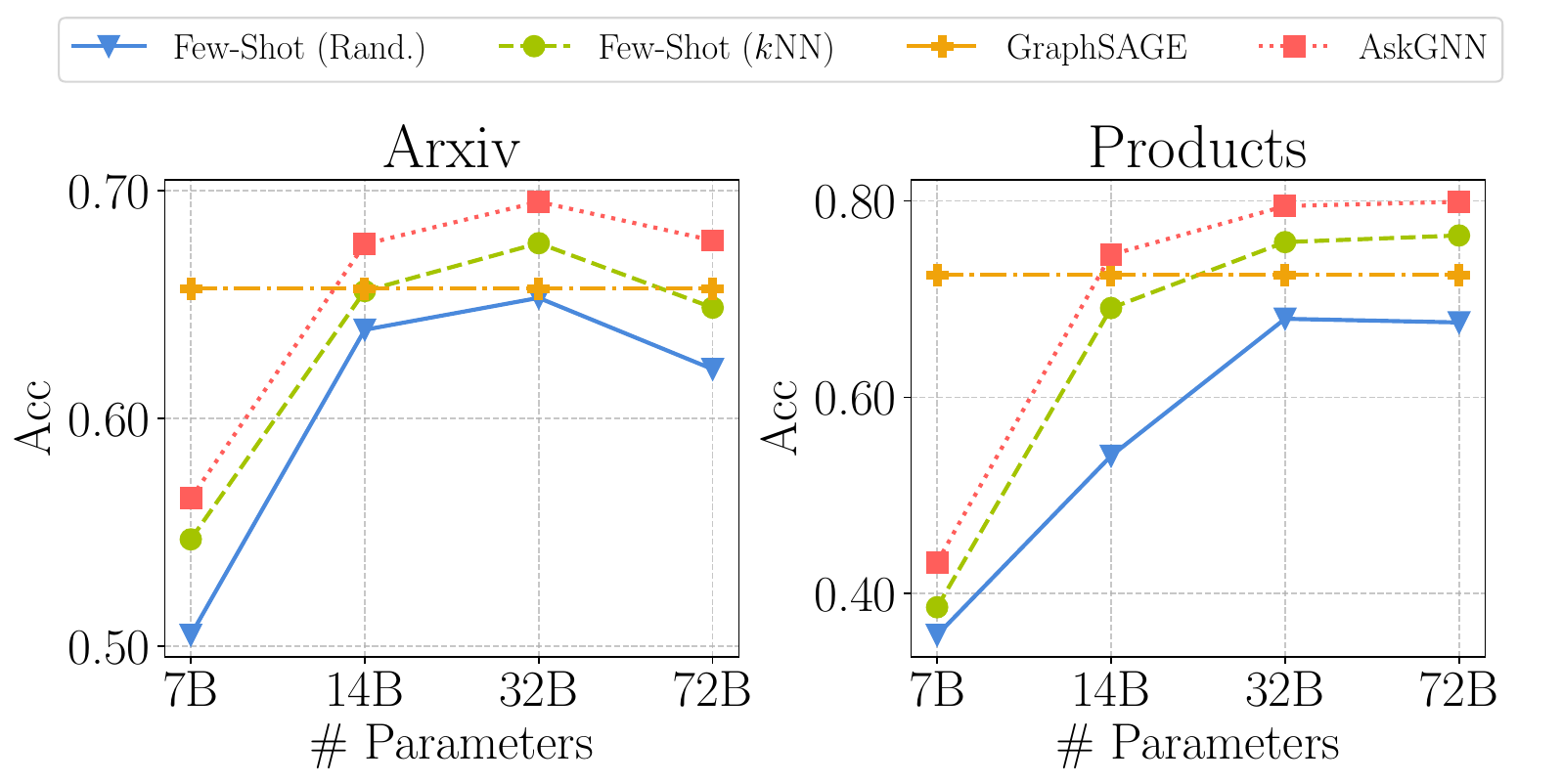}
\caption{
Performance of the Qwen1.5 family across parameter sizes (7B to 72B) on ogbn-product and ogbn-arxiv datasets. 
}
\label{fig:sameArc}
\end{figure}

As shown in~\autoref{fig:sameArc}, we explore the scalability of the Qwen1.5~\citep{bai2023qwen} architecture on the ogbn-product and ogbn-arxiv datasets, with model scales ranging from 7B to 72B parameters.
The empirical results indicate that \OURS{} consistently outperforms both few-shot ($k$-NN) and few-shot (Rand.), highlighting the effectiveness of selecting structure ICL examples across all model scales.
Interestingly, we observe an inverse scaling phenomenon, where performance initially rises but then declines as the model size increases from 7B to 72B parameters.
This behavior, seen in both datasets, contradicts the conventional expectation that larger models should inherently yield better performance. 
For ogbn-product, the 72B model shows lower accuracy compared to the 32B model, and a similar trend is observed for ogbn-arxiv
This aligns with previous findings~\citep{mckenzie2023inverse, dohmatob2024tale}, which suggest that inverse scaling behavior arises from misleading few-shot demonstrations, where larger models capture nuanced language aspects but are also misled by noise or non-representative features.

\begin{figure}[h]
    \includegraphics[width=\linewidth]{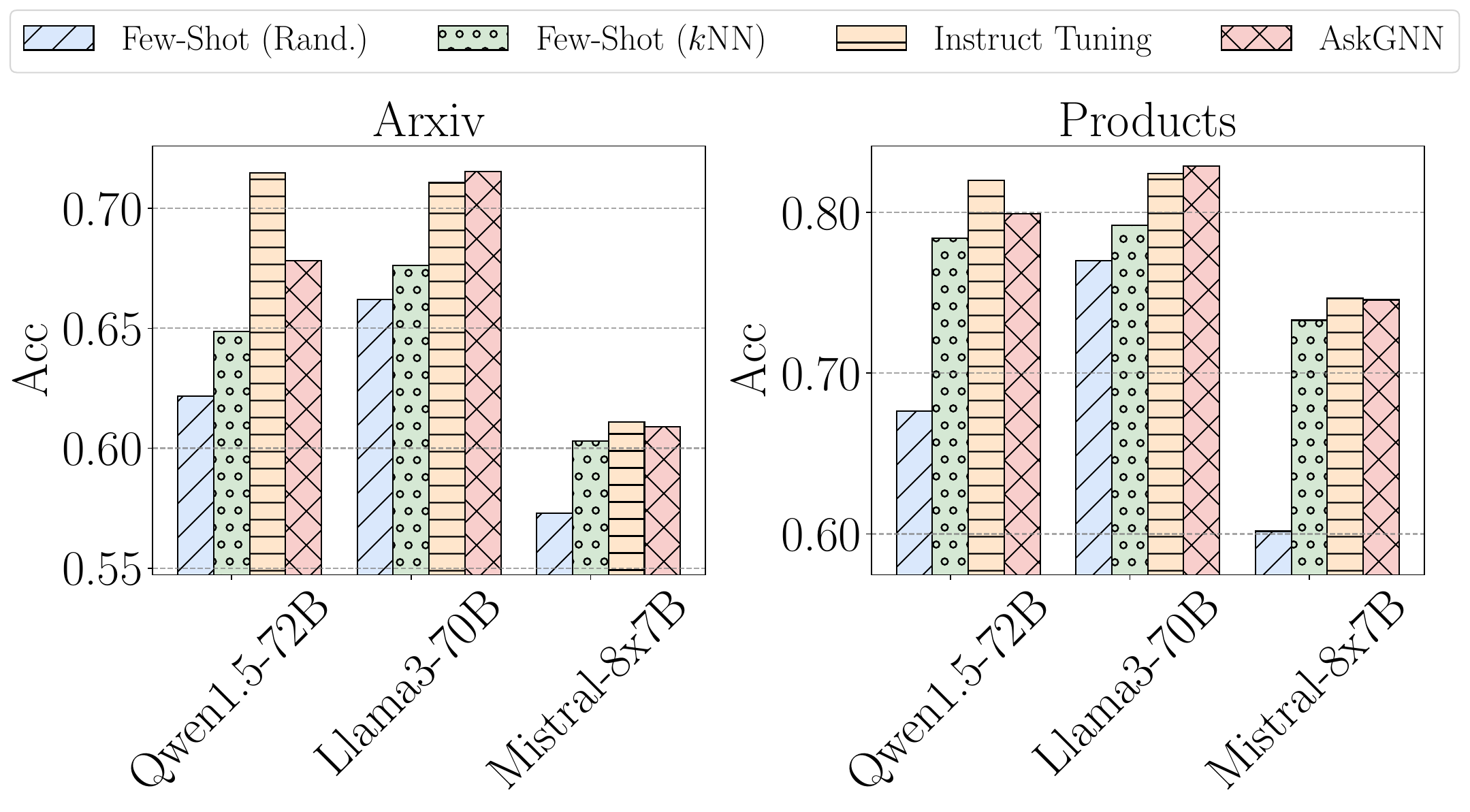}
\caption{
Performance comparison of \OURS{} across different LLM architectures, including Dense models (Qwen1.5-72B, Llama3-70B) and MoE models (Mistral-8x7B). 
}
\label{fig:diffArc}
\end{figure}

\paragraph{Different LLM Families.}
We tested our framework across various LLM families, including Dense models like Qwen1.5-72B and Llama3-70B, as well as  Mixture of Experts (MoE) models like Mistral-8x7B. 
The results are shown in \autoref{fig:diffArc}.
Our results show that {\OURS} consistently outperforms Few-Shot, emphasizing the importance of selecting ICL examples across different model architectures, whether Dense or MoE. 
Notably, for less powerful models, optimal ICL example selection leads to greater performance gains, with a larger gap between \OURS{} and Few-shot ($k$-NN) compared to Few-shot (Rand.). This highlights the significant impact of example selection on enhancing the capabilities of smaller or weaker models. 
For weaker models (e.g., Qwen1.5-72B), instruct tuning surpasses {\OURS}. However, for stronger models like Llama3-70B, {\OURS} proves more effective, likely due to Llama3-70B's training on more relevant tasks.
Additionally, the performance gap between instruct tuning and {\OURS} narrows under the MoE architecture. This is likely due to MoE's dynamic allocation capabilities, which enhance flexibility and efficiency in processing diverse inputs, leading to better integration of instruct tuning and ICL, thereby reducing the disparity between the two methods.
\begin{table}
\caption{
Comparison of performance between Majority Voting (MV), Few-shot ($k$-NN), and \OURS{} across various experimental settings. MV selects the most frequent class in the ICL examples and consistently underperforms compared to LLM predictions, demonstrating the LLM's ability to make more nuanced decisions beyond simple class frequency. 
}
\scalebox{0.7}{
\begin{tabular}{@{}lcccc@{}}
\toprule
\multicolumn{1}{c}{\multirow{2}{*}{\textbf{Methods}}} & \multicolumn{4}{c}{\textbf{ogbn-arxiv}}               \\ \cmidrule(l){2-5} 
\multicolumn{1}{c}{}                                  & 1\%         & 3\%         & 5\%         & 10\%        \\ \midrule
Few-Shot ($k$-NN)                                              & 61.90\scriptsize$\pm$ 0.02  & 62.06\scriptsize$\pm$ 0.11  & 62.30\scriptsize$\pm$ 0.14  & 62.17\scriptsize$\pm$ 0.31  \\
MV+$k$-NN & 29.97 \scriptsize$\pm$0.32        & 35.50 \scriptsize$\pm$0.82        & 34.10 \scriptsize$\pm$0.10        & 35.33 \scriptsize$\pm$0.12   \\ \midrule
{\OURS}                                                 & 69.13\scriptsize$\pm$ 0.24  & 69.42\scriptsize$\pm$ 0.12  & 70.29\scriptsize$\pm$ 0.25  & 71.53\scriptsize$\pm$ 0.11  \\
MV+{\OURS} & 57.27\scriptsize$\pm$ 0.21  & 59.37\scriptsize$\pm$ 0.17  & 58.00\scriptsize$\pm$ 0.10  & 60.87\scriptsize$\pm$ 0.40  \\ \bottomrule

& \multicolumn{4}{c}{\textbf{ogbn-products}}            \\ \cmidrule(l){2-5} 
\multicolumn{1}{c}{}                                  & 1\%         & 3\%         & 5\%         & 10\%        \\ \midrule
Few-Shot ($k$-NN)                                             & 66.80\scriptsize$\pm$ 0.04 & 67.38\scriptsize$\pm$ 0.19 & 67.18\scriptsize$\pm$ 0.27 & 67.62\scriptsize$\pm$ 0.08  \\
MV + $k$-NN                                         & 47.30 \scriptsize$\pm$0.44        & 54.69 \scriptsize$\pm$0.31        & 54.09 \scriptsize$\pm$0.72        & 53.01 \scriptsize$\pm$0.31  \\ \midrule
{\OURS}                                           & 65.24\scriptsize$\pm$ 0.20 & 68.24\scriptsize$\pm$ 0.19 & 82.54\scriptsize$\pm$ 0.13 & 82.88\scriptsize$\pm$ 0.17  \\
MV + {\OURS}                                            & 65.24\scriptsize$\pm$ 0.20 & 68.24\scriptsize$\pm$ 0.21 & 71.39\scriptsize$\pm$ 0.39 & 71.63\scriptsize$\pm$ 0.31  \\ \bottomrule
\end{tabular}
}

\label{tab:t1}
\end{table}

\begin{table*}[h]
\centering
% \small
\caption{Experiment results of different heuristic ICL examples retrieve appraoches. 
The best results are \textbf{bold}.
}
\scalebox{0.69}{
  
\begin{tabular}{cc lll lll lll lll c}
\toprule
& &  \multirow{2}{*}{\textbf{Methods}} & \multicolumn{3}{c}{\textbf{ogbn-arxiv}}                                                                       & \multicolumn{3}{c}{\textbf{ogbn-products}}       & \multicolumn{3}{c}{\textbf{Arxiv2023}}                                                              & \multirow{2}{*}{\textbf{Avg.}} \\ \cmidrule(l){4-12} 
& &  & \multicolumn{1}{c}{1\%}     & \multicolumn{1}{c}{5\%}   & \multicolumn{1}{c}{10\%}  & \multicolumn{1}{c}{1\%}   & \multicolumn{1}{c}{5\%}   & \multicolumn{1}{c}{10\%}  & \multicolumn{1}{c}{1\%}     & \multicolumn{1}{c}{5\%}   & \multicolumn{1}{c}{10\%}   &    \\ \midrule
 \multirow{3}{*}{\STAB{\rotatebox[origin=c]{90}{{Qwen1.5}}}} & \multirow{3}{*}{\STAB{\rotatebox[origin=c]{90}{{72B}}}}
      & \textit{NPL}                                   & {{63.91}} \scriptsize$\pm$0.26  & 63.91 \scriptsize$\pm$0.26 & 63.91 \scriptsize$\pm$0.26 & 73.83 \scriptsize$\pm$0.17  & 73.83 \scriptsize$\pm$0.17 & 73.83 \scriptsize$\pm$0.17  & 63.07 \scriptsize$\pm$0.83                                     & 65.78 \scriptsize$\pm$0.69                 & 65.82 \scriptsize$\pm$0.43     & 67.54 \\
     &  & \textit{NPG}                                   & 62.43 \scriptsize$\pm$0.38  & 66.47 \scriptsize$\pm$0.29 & {{65.90}} \scriptsize$\pm$0.33 & 75.23 \scriptsize$\pm$0.34 & 76.62 \scriptsize$\pm$0.26 & 77.89 \scriptsize$\pm$0.17 & 62.17 \scriptsize$\pm$0.23                                     & 63.46 \scriptsize$\pm$0.17                 & 64.62 \scriptsize$\pm$0.13   &  68.42 \\ 
  \cmidrule{3-13}
\rowcolor[HTML]{EFEFEF}  & & {{\textbf{\OURS{}}}}                                       & {\textbf{64.67}} \scriptsize$\pm$0.10  & {\textbf{66.95}} \scriptsize$\pm$0.23 & {\textbf{67.82}} \scriptsize$\pm$0.16 & {\textbf{76.79}} \scriptsize$\pm$0.26 & {\textbf{78.72}} \scriptsize$\pm$0.19 & {\textbf{79.91}} \scriptsize$\pm$0.23   & {\textbf{69.56}} \scriptsize$\pm$0.22        & {\textbf{69.97}} \scriptsize$\pm$0.12        & \multicolumn{1}{c}{{\textbf{70.46}} \scriptsize$\pm$0.17}  & {{71.65}} \\ \midrule

 \multirow{3}{*}{\STAB{\rotatebox[origin=c]{90}{{Llama3}}}} & \multirow{3}{*}{\STAB{\rotatebox[origin=c]{90}{{70B}}}}
  & \textit{NPL}                                   & \textbf{69.49} \scriptsize$\pm$0.35          & {69.49} \scriptsize$\pm$0.35        & 69.49 \scriptsize$\pm$0.35        & 80.86 \scriptsize$\pm$0.18           & 80.86 \scriptsize$\pm$0.18        & 80.86 \scriptsize$\pm$0.18      & 72.27 \scriptsize$\pm$0.19  & 72.27 \scriptsize$\pm$0.19  & 72.27 \scriptsize$\pm$0.19  &   73.65      \\
  & & \textit{NPG}                                   & 65.50 \scriptsize$\pm$0.20         & 68.57 \scriptsize$\pm$0.21        & 68.33 \scriptsize$\pm$0.21        & 77.61 \scriptsize$\pm$0.06                & 79.03 \scriptsize$\pm$0.16        & 80.22 \scriptsize$\pm$0.27      & 67.77 \scriptsize$\pm$0.22  & 70.35 \scriptsize$\pm$0.14  & 71.08 \scriptsize$\pm$0.19    &  71.16        \\
  \cmidrule{3-13}
 \rowcolor[HTML]{EFEFEF} & & {\textbf{\OURS{}}}                                       & {{69.13}} \scriptsize$\pm$0.24             & {\textbf{70.29}} \scriptsize$\pm$0.25        & {\textbf{71.53}} \scriptsize$\pm$0.11        & {\textbf{81.35}} \scriptsize$\pm$0.23              & {\textbf{82.54}} \scriptsize$\pm$0.13        & {\textbf{82.88}} \scriptsize$\pm$0.17      & {\textbf{73.86}} \scriptsize$\pm$0.22        & {\textbf{74.07}} \scriptsize$\pm$0.12        & \multicolumn{1}{c}{{\textbf{74.97}} \scriptsize$\pm$0.17}  &  \textbf{{75.62}} \\
\bottomrule
                                                          
\end{tabular}
}

\label{tab:varients}
\end{table*}

\begin{figure}[h]
\centering
\includegraphics[width=1.0\linewidth]{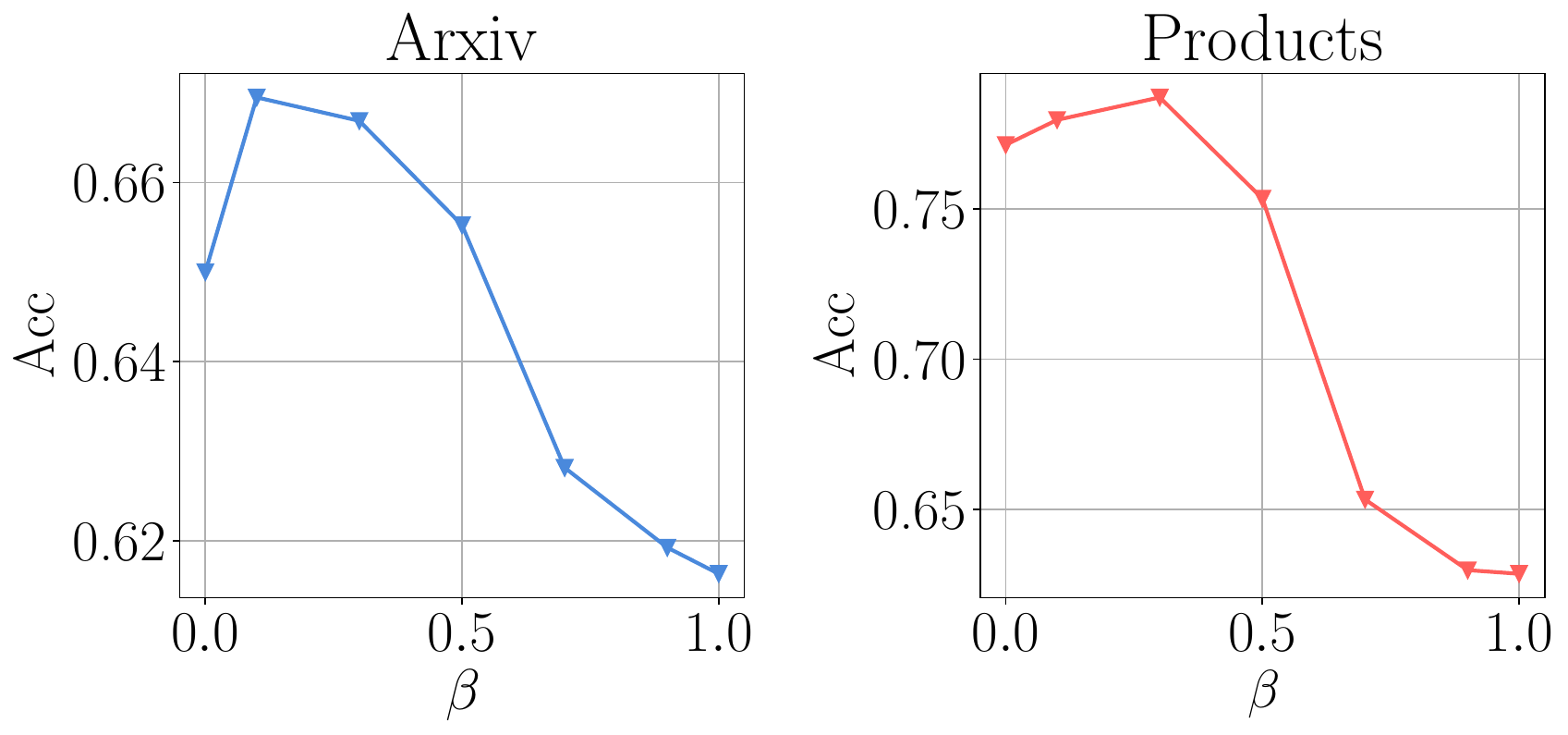}
\caption{
Performance variation with respect to hyperparameter $\beta$ on \texttt{ogbn-arxiv} and \texttt{ogbn-products} datasets using Qwen1.5-72B.}
    \label{fig:Ablationbeta}
\end{figure}

\subsection{Hyperparameter Analysis on $\beta$}
To investigate the impact of the classification loss $\mathcal{L}_{\text{clf}}$ introduced in Equation \ref{eq:finalloss}, we conducted a comprehensive hyperparameter analysis on the weighting factor $\beta$. \autoref{fig:Ablationbeta} illustrates the performance trends as $\beta$ varies from 0.0 to 1.0 on both the \texttt{ogbn-arxiv} and \texttt{ogbn-products} datasets using the Qwen1.5-72B model. For \texttt{ogbn-arxiv}, we observe peak performance at $\beta \approx 0.2$, suggesting a balanced contribution from both feedback-based and structural-based losses maximizes performance. In contrast, \texttt{ogbn-products} exhibits higher sensitivity to $\beta$, peaking at $\beta = 0.3$ with a steeper subsequent decline, indicating a greater dependence on the structural component. We hypothesize this is due to the complex nature of product descriptions benefiting more from structural insights.
These observations underscore the importance of fine-tuning the balance between different components of the loss function tailored to the specific characteristics of the dataset, thereby enabling optimal model performance.
These observations underscore the importance of fine-tuning the balance between different components of the loss function tailored to the specific characteristics of the dataset, thereby enabling optimal model performance.

\subsection{Analysis on ICL Examples}
\paragraph{Cast Study.} We provide a case study of retrieved examples from the ogbn-product dataset to illustrate the difference in example selection between \OURS{} and $k$-NN, as shown in \autoref{tab:case_study}. 
This case study demonstrates that while $k$-NN selects semantically similar examples (focusing on underwater themes), it misses the crucial ``Toy'' classification. In contrast, {\OURS} retrieves an example that, despite having lower semantic similarity, correctly captures the ``Toy'' category. This highlights {\OURS}'s ability to prioritize task-relevant information over semantic similarity, resulting in more accurate classifications for the LLM.
\begin{table}[tbh!]
    \caption{
    Comparison of retrieved examples from the ogbn-product dataset. \textbf{Bold} words indicate semantic similarity.    }
    \centering
    \small
    \begin{tabular}{cLc}
    \toprule
    Type & Text & Label \\
    \midrule
    Query & wonder \textbf{pets flyboat} with 3 removable figures for everyday fun	& Toy \\
    \midrule
    $k$-NN & dive into adventure with your favorite \textbf{underwater} explorers, the \textbf{octonauts}! on this exciting dvd, captain barnacles tangles with a colossal squid	& Movie
    \\
    \midrule
    {\OURS} & choose your favorite paw patrol character and rush into adventure bay action with their special vehicle!	& Toy\\
    \bottomrule
    
    \end{tabular}
\label{tab:case_study}
\end{table}

\paragraph{Comparison with 1-hop Neighbors with Pseudo Labels.}
\label{sec:ablation}
To understand the importance of retrieve ICL examples across whole graph, we consider utilize the query node's 1-hop neighbors. \citet{huang2023can} suggests that leveraging a query's neighbor and its label can yield satisfactory results. 
However, since actual labels for the query nodes' neighbors are not accessible, 
we design two approaches for getting the pseudo label: NPL and NPG, 
NPL stands for \underline{N}eighbors \underline{P}seudo \underline{L}abels generated by the \textbf{L}LM, while NPG stands for \underline{N}eighbors' \underline{P}seudo \underline{}{L}abels generated by the \underline{G}NN.
Results in \autoref{tab:varients} show that \OURS{} surpasses \textit{GE}, \textit{NPL}, and \textit{NPG}, highlighting the importance of retrieving high-quality samples for LLM prediction. 
Notably, even with pseudo labels, \textit{NPL} and \textit{NPG} outperform \textit{GE} with true labels, suggesting that effective ICL examples can significantly improve LLM predictions, even when the labels are noisy.
Finally, \textit{NPG} outperforms \textit{NPL} with Qwen1.5-72B, while \textit{NPL} performs better than \textit{NPG} with Llama3-70B.
Combined with the GNN and Zero-Shot results in \autoref{tab:nc_result}, we conclude that the quality of the pseudo-label is crucial for effectively using a query node's neighbors as ICL examples.

\begin{figure}[h]
    \centering
    \includegraphics[width=0.95\linewidth]{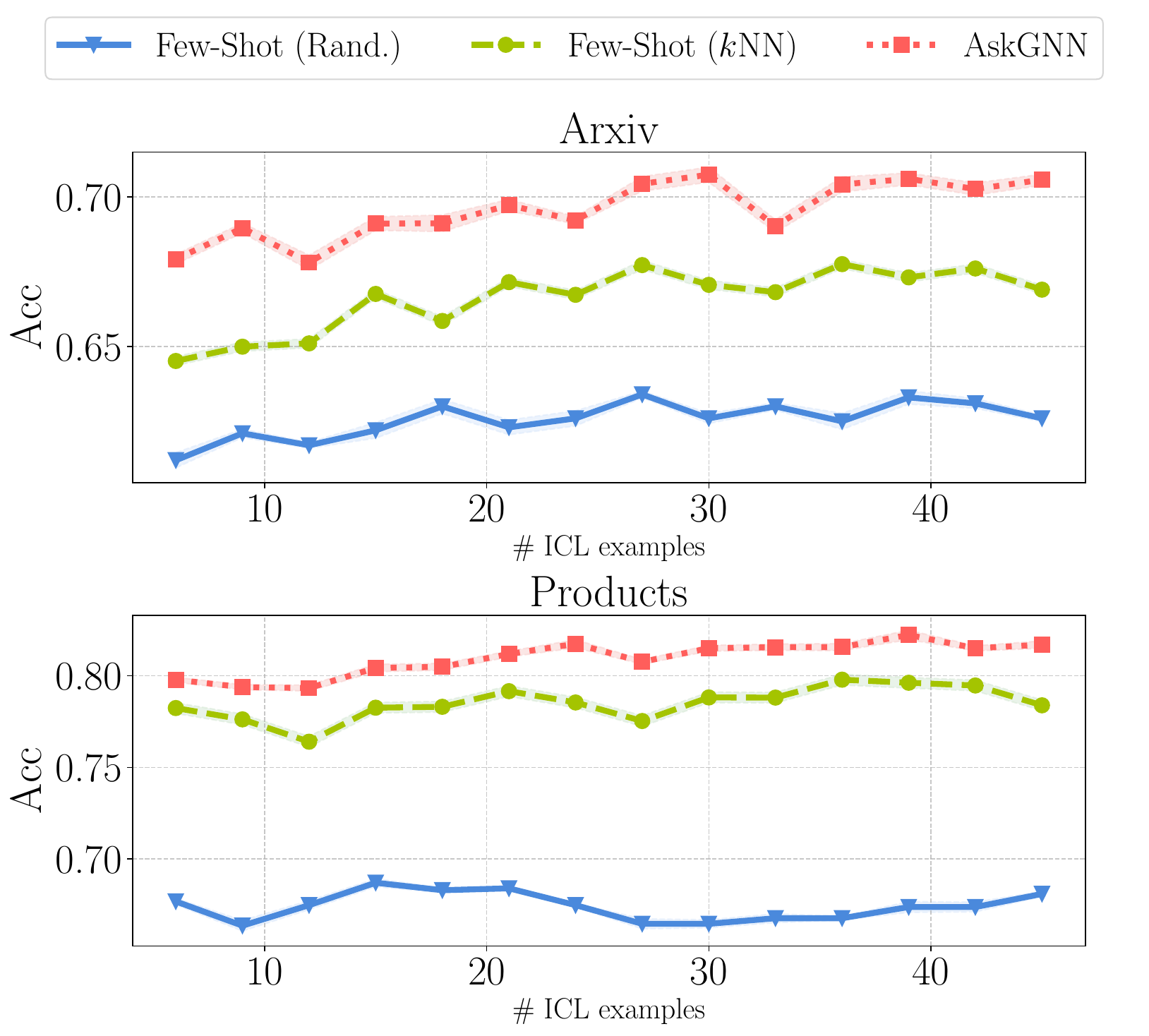}
\caption{
Performance comparison of \OURS{}, Few-shot (Rand.), and Few-shot ($k$-NN) with varying numbers of ICL examples on the ogbn-arxiv and ogbn-products datasets. 
}
    \label{fig:dataefficient}
\end{figure}

\paragraph{\# of ICL Examples.}
\label{sec:EffiofBS}
To evaluate the generalization ability of our approach across different numbers of ICL examples, we conducted an experiment to assess its performance under varying conditions.
This analysis is essential for understanding the robustness and scalability of {\OURS} across different datasets.
As shown in \autoref{fig:dataefficient}, we compare the sample efficiency of  Few-shot (Rand.), Few-shot ($k$-NN) and \OURS{}. 
We observe that performance improves for all methods as the number of ICL examples increases.
This improvement is more pronounced on the ogbn-arxiv dataset than on the ogbn-products dataset.
Additionally, our method consistently outperforms Few-Shot across all datasets, demonstrating its strong generalization capability, even as the ratio of ICL examples increases.

\paragraph{ICL Example Purification.}
\label{sec: NoiseReduction}
To address the potential impact of noise in retrieved In-Context Learning (ICL) examples, we conducted experiments exploring two noise reduction strategies through ICL example purification. Given that performance often plateaus as context length increases \citep{li2024long}, we focused on two heuristic approaches: \textit{(1).} LLM-Selection. This approach involves prompting the Large Language Model (LLM) to select the most informative examples from multiple retrieved instances based on criteria such as relevance and diversity. The underlying intuition is that LLMs, when tasked with curating their own input data, can identify examples that offer the most value for downstream tasks while filtering out noisy or redundant entries;
\textit{(2).} Minority-Class-Removal. This method involves removing ICL examples with less frequent labels. We hypothesized that such outliers might introduce unwanted noise into the retrieval process. By removing these minority class examples, we aimed to reduce the likelihood of including anomalous data that could detract from model performance.

The results from these experiments, conducted using Qwen1.5-72B on the obgn-arxiv dataset, are summarized in~\autoref{tab:Noise}.
Our findings indicate that LLM-Selection improved performance by approximately 1\% (e.g., from 69.02\% to 71.10\% accuracy on ogbn-arxiv). In contrast, Minority-Class-Removal had a minimal impact, with changes in performance generally within ±0.1\%. 
Overall, these additional experiments demonstrate that improving the quality of retrieved ICL examples can positively impact performance.

\subsection{Are LLMs Just Repeaters?}
\label{sec:LLMReader}
To determine whether LLMs merely replicate the most prevalent class from ICL examples, rather than engaging in meaningful graph reasoning and task understanding, we analyzed performance in two scenarios: Few-shot ($k$-NN) and \OURS{}.
Additionally, we implemented a Majority Voting (MV) strategy, which selects the class that appears most frequently in the ICL examples as the prediction.
As shown in \autoref{tab:t1}, MV consistently underperformed compared to LLM predictions across all experimental conditions, indicating that LLMs are making more nuanced decisions rather than simply mimicking the most common class.
This outcome highlights the importance of selecting relevant examples strategically in ICL.

\begin{table}
\caption{Performance of ICL example purification techniques on the ogbn-arxiv dataset using Qwen1.5-72B. 
}
\small
\scalebox{0.76}{
\begin{tabular}{@{}lcccccc@{}}
\toprule
& \multicolumn{3}{c}{\textbf{ogbn-arxiv}} & \multicolumn{3}{c}{\textbf{ogbn-products}} \\ \midrule
{\color[HTML]{333333} \textbf{\# of ICL Examples}}                                                                   & 30          & 33          & 36          & 24           & 27           & 30           \\ \midrule
{\color[HTML]{333333} {\OURS{}}}                                                                            & 70.74       & 69.02       & 70.43       & 81.44        & 80.76        & 81.50        \\ \cmidrule(l){2-7} 
 LLM Selection& 70.81       & 71.10       & 71.39       & 81.49        & 81.66        & 81.70        \\ \cmidrule(l){2-7} 
Minority-Class-Removal & 70.72       & 69.17       & 70.41       & 81.42        & 80.77        & 81.52        \\ \bottomrule
\end{tabular}
}

\label{tab:Noise}
\end{table}

\section{Conclusion}
\vspace{-1mm}
% \vspace{-2mm}
In this work, we introduced \OURS{}, a novel approach leveraging the ICL capabilities of LLMs for graph-based tasks. 
By implementing a structure-enhanced retriever (SE-Retriever)  based on GNNs, \OURS{} bridges the gap between sequential text processing and graph-structured data while preserving the inherent supervision signals of graph nodes. 
Our evaluations across three distinct tasks and seven LLMs show that \OURS{} significantly improves node classification performance compared to existing methods, confirming its robustness and effectiveness. 
Looking forward, future work will focus on extending \OURS{} to a broader range of graph-based applications, such as  dynamic graph analysis, where the complexity of the data increases and requires more sophisticated handling.
This research establishes a promising foundation for utilizing structured graph data to expand the capabilities of LLMs in advanced applications.

\section{Limitations}
\vspace{-1mm}
We also recognize the following limitations of this work: First, data quality issues, such as low-quality product descriptions, can hinder larger models more than smaller ones, affecting performance on complex datasets. Second, the limited input window of current open-source LLMs capped the number of ICL examples at 45, restricting the exploration of larger example sets that might improve performance. Last, our approach relies heavily on the learned GNN retriever, meaning inaccurate structure information could reduce overall model effectiveness.

\section{Acknowledgement}
\vspace{-1.5mm}
This research was supported in part by NIH R01LM01372201.

\clearpage

\bibliography{main}

\clearpage

\appendix

\section{Baseline Details.} 
\label{sec:baselineDetail}

\begin{itemize}[leftmargin=*, itemsep=0.5pt, topsep=2.5pt]

    \item \textbf{Bare GNN}: 
    \textit{GCN}~\citep{ye2024language} is a semi-supervised learning approach for graph-structured data, employing convolutional operations to aggregate and transform node features. It is particularly effective in tasks such as node classification and link prediction.
    \textit{GraphSAGE}~\citep{ye2024language} is an inductive framework for learning node embeddings by sampling and aggregating features from a node's neighbors, thereby enabling scalability for large graph datasets.
    \textit{GraphSAINT}~\citep{ye2024language} enhances the scalability of GNNs through graph sampling, reducing computational burden and memory usage by generating smaller subgraphs for training, making it well-suited for large-scale graph tasks.

    \item \textbf{Text-based Serialization}: 
    \textit{K-Hop}~\citep{chen2024exploring} leverages 2-hop neighbors as in-context examples, utilizing the typical 2-layer structure of GNNs for enhanced context.
    \textit{Graph-CoT}~\citep{jin2024graph} augments LLMs through iterative reasoning on graphs via the Graph Chain-of-Thought framework, consistently outperforming baseline methods.

    \item \textbf{Graph Projection}: 
    \textit{InstructGLM}~\citep{ye2024language} employs natural language prompts to describe graph structures, enabling generative language models to handle graph machine learning tasks. This approach eliminates the need for complex graph-specific mechanisms and supports scalable training.

    \item \textbf{InstructTuning}: 
    This technique fine-tunes LLM parameters to enable the model to learn information within graph datasets~\citep{zheng2024llamafactory}.

    \item \textbf{In-Context Learning}: 
    \textit{Zero-Shot}~\citep{chen2023label} relies solely on the node's attributes for prediction without additional context.
    \textit{Few-Shot (Rand.)}~\citep{chen2023label} involves randomly selecting a few labeled examples along with the node's attributes to assist LLMs in understanding the task, utilizing random sampling for these selections.
    \textit{Few-Shot ($k$-NN)}~\citep{lewis2020retrieval} selects the most similar sample to the query as the in-context example based on retrieval-augmented generation.

\end{itemize}

\section{Prompt Design}
\label{sec: prompt design}

\autoref{tab: prompt} presents the prompts utilized across different datasets. Each prompt begins with the abstract and title of the paper, followed by a task-specific question designed to probe the model on a particular aspect of the paper and to request an explanation for the prediction. The answer section is left blank for the model to complete. Our analysis indicates that the current instructions enable the LLM to generate output that closely adheres to the expected format, with minimal deviations. This consistency facilitates the straightforward extraction of answers from the LLM's text output.

\begin{table*}[h]
\small
    \centering
    % \begin{tabular}{lcc}
   \caption{Prompts used in this study to query the LLM.}
    \label{tab: prompt}
    \begin{tabularx}{\textwidth}{lX}
    \toprule
         Dataset & Prompt\\
         \midrule
         \texttt{ogbn-arxiv \& arxiv23}

         &{ You are an AI trained to categorize arxiv computer science papers into specific categories based on their abstracts. Your task is to analyze the paper description provided and identify the most relevant category. \newl Paper description:\promptfield{paper description}. \newl Give me the category of this content. Respond only with the category key (e.g., 'cs.AI', 'cs.SY'), without any additional text or explanation. Here are some of the papers cited by this paper: \promptfield{ICL examples}: \newl \newl Answer:}\\

         \midrule
         \texttt{ogbn-products}
          &{ You are an AI trained to categorize products into specific categories based on their descriptions and characteristics. Your task is to analyze the product description provided, consider its characteristics, and identify the most relevant category among hundreds of possible categories. There are a total of 46 categories, including  1) Home \& Kitchen, 2) Health \& Personal Care, 3) Beauty, 4) Sports \& Outdoors, 5) Books, 6) Patio, Lawn \& Garden, 7) Toys \& Games, 8) CDs \& Vinyl, 9) Cell Phones \& Accessories, 10) Grocery \& Gourmet Food, 11) Arts, Crafts \& Sewing, 12) Clothing, Shoes \& Jewelry, 13) Electronics, 14) Movies \& TV, 15) Software, 16) Video Games, 17) Automotive, 18) Pet Supplies, 19) Office Products, 20) Industrial \& Scientific, 21) Musical Instruments, 22) Tools \& Home Improvement, 23) Magazine Subscriptions, 24) Baby Products, 25) NAN, 26) Appliances, 27) Kitchen \& Dining, 28) Collectibles \& Fine Art, 29) All Beauty, 30) Luxury Beauty, 31) Amazon Fashion, 32) Computers, 33) All Electronics, 34) Purchase Circles, 35) MP3 Players \& Accessories, 36) Gift Cards, 37) Office \& School Supplies, 38) Home Improvement, 39) Camera \& Photo, 40) GPS \& Navigation, 41) Digital Music, 42) Car Electronics, 43) Baby, 44) Kindle Store, 45) Kindle Apps, 46) Furniture. \newl  Product description: {\color{blue}<product description>}. \newl Consider its characteristics and give me the category of this product. Respond only with the category key (e.g., 'Electronics', 'Toys \& Games'), without any additional text or explanation. \newl Here are some examples to help you understand how to categorize products based on their descriptions: \promptfield{ICL examples}.   \newl\newl Answer:}

         \\

         \bottomrule
    \end{tabularx}
\end{table*}

\paragraph{Exploring Prompt Variations.}

We conducted an extensive exploration of the impact of different prompts on the \texttt{ogbn-arxiv} dataset.
As shown in \autoref{tab: prompt experiment}, the performance across most prompts is generally similar. However, a slight improvement in accuracy is noted when the title is placed after the abstract. This observation supports the principle suggested by \cite{zhao2021calibrate}, which posits that positioning more critical information later in the prompt can enhance performance.

\begin{table*}[h]
\small
    \centering
    % \begin{tabular}{lcc}
    \caption{Prompts employed in our experiments to investigate the impact of various prompting strategies. The results indicate relatively uniform performance across most prompts.}
    \label{tab: prompt experiment}
    \begin{tabularx}{\textwidth}{lXc}
    \toprule
         Description & Prompt & Accuracy\\
         \midrule
         Default prompt & You are an AI trained to categorize arxiv computer science papers into specific categories based on their abstracts. Your task is to analyze the paper description provided and identify the most relevant category. \newl Paper description: Abstract: \promptfield{abstract text} \newl Title: \promptfield{title text}. \newl Give me the category of this content. Respond only with the category key (e.g., 'cs.AI', 'cs.SY'), without any additional text or explanation. Here are some of the papers cited by this paper: \promptfield{ICL examples}: \newl \newl Answer: & 0.729\\  
         \midrule
         Title first & You are an AI trained to categorize arxiv computer science papers into specific categories based on their titles and abstracts. Your task is to analyze the paper description provided and identify the most relevant category. \newl \textbf{Title:} \promptfield{title text} \newl Paper description:\promptfield{paper description}. \newl Give me the category of this content. Respond only with the category key (e.g., 'cs.AI', 'cs.SY'), without any additional text or explanation. Here are some of the papers cited by this paper: \promptfield{ICL examples}: \newl \newl Answer: & 0.700\\  
         \midrule
         Focus on text content & You are an AI trained to categorize arxiv computer science papers into specific categories based on their abstracts. Your task is to analyze the paper description provided and identify the most relevant category. \newl Paper description:\promptfield{paper description}. \newl Give me the category of this content. \textbf{Focus only on content in the actual text and avoid making false associations.} Respond only with the category key (e.g., 'cs.AI', 'cs.SY'), without any additional text or explanation. Here are some of the papers cited by this paper: \promptfield{ICL examples}: \newl \newl Answer: & 0.698\\  
         \midrule
         Chain of thought prompt & You are an AI trained to categorize arxiv computer science papers into specific categories based on their abstracts. Your task is to analyze the paper description provided and identify the most relevant category. \newl Paper description: Abstract: \promptfield{abstract text} \newl Title: \promptfield{title text}. \newl Give me the category of this content. \textbf{Please think about the categorization in a step by step manner and avoid making false associations.} Respond only with the category key (e.g., 'cs.AI', 'cs.SY'), without any additional text or explanation. Here are some of the papers cited by this paper: \promptfield{ICL examples}: \newl \newl Answer: & 0.709\\  
         \bottomrule
    \end{tabularx}
\end{table*}

\begin{table}[h]

    \caption{Statistics of the TAG datasets}
    \label{tab: dataset}
    \scalebox{0.75}{
    \begin{tabular}{lccc}
    \toprule
         Dataset &  \#Nodes &\#Edges & Task   \\
         \midrule

         \texttt{ogbn-arxiv} & 169,343 & 1,166,243 & 40-class classif.
         \\
         \texttt{ogbn-products}  & 2,449,029	&61,859,140	 & 47-class classif.  \\
         \texttt{arxiv23}
         & {46,198}
         & {78,548}
         &{40-class-classif.}
         
         \\
         \bottomrule
    \end{tabular}
    }
\end{table}

\section{Dataset}

\label{sec:dataset}

We conduct experiments on three TAGs: 
\texttt{ogbn-arxiv}, \texttt{ogbn-products}~\citep{ogb-datasets}, and {arxiv23}. \autoref{tab: dataset} summarizes the statistics of these datasets.

\subsection{Dataset Description}

\begin{itemize}
    \item \textbf{ogbn-arxiv~\citep{ogb-datasets}.}
    The \texttt{ogbn-arxiv} dataset is a directed graph representing the citation network of all computer science arxiv papers indexed by MAG~\citep{wang2020microsoft_mag}. Each node is an arxiv paper, and each directed edge represents a citation. The task is to predict the 40 subject areas of arxiv CS papers, such as cs.AI, cs.LG, and cs.OS, which are manually labeled by the authors and arxiv moderators.
    
    \item \textbf{ogbn-products~\citep{ogb-datasets}.}
    The \texttt{ogbn-products} dataset represents an Amazon product co-purchasing network, with nodes representing products and edges indicating co-purchases. The task is to predict the category of a product in a multi-class classification setup, using 47 top-level categories as target labels.
    
    \item \textbf{arxiv23~\citep{he2023harnessing}}. 
    The \texttt{arxiv23} dataset is a directed graph representing the citation network of all computer science arxiv papers published in 2023 or later. Similar to \texttt{ogbn-arxiv}, each node is an arxiv paper, and each directed edge represents a citation. The task is to predict the 40 subject areas of arxiv CS papers, such as cs.AI, cs.LG, and cs.OS, which are manually labeled by the authors and arxiv moderators.
\end{itemize}

\subsection{Dataset splits and random seeds}
In our experiments, we adhered to specific dataset splits and employed random seeds for reproducibility. For the \texttt{ogbn-arxiv} and \texttt{ogbn-products} datasets, we used the standard train/validation/test split provided by OGB~\citep{ogb-datasets}. For  \texttt{arxiv23} datasets, we followed the splits used in previous works~\citep{huang2023can,he2023harnessing}. Additionally, we utilized 10 random seeds to ensure the reproducibility of our experiments. This approach enabled consistent evaluation of our proposed method across the respective datasets, with detailed results available in the supplementary material.

\section{Experiment Details}
\label{app sec: experiment}

\paragraph{Computing Environment and Resources.}
The proposed method was implemented using PyG modules, which are licensed under the MIT License. Our experiments were conducted on a high-performance computing setup featuring an Intel(R) Xeon(R) Platinum 8358P CPU at 2.60GHz, with 512GB of memory. The computational resources included eight NVIDIA A6000 GPUs, each with 48GB of memory.
\begin{table}[h]
    \caption{Hyperparameters for the GCN, GraphSAGE, and GraphSAINT models.}
    \label{tab: hyperparam}
    \scalebox{0.85}{
    \begin{tabular}{lccc}
    \toprule
Hyperparameters & GCN& GraphSAGE&GraphSAINT\\
\midrule
\# layers & 3 & 3 & 3\\
hidden dim & 256 & 256 & 256 \\
learning rate & 0.01 & 0.01 & 0.001\\
dropout & 0.5 & 0.5 & 0.5\\
epoch & 200 & 200 & 200\\

\bottomrule
\end{tabular}
}
\end{table}

\paragraph{Hyperparameters.}
\autoref{tab: hyperparam} presents the hyperparameters utilized for the GCN~\citep{kipf2017semisupervised}, GraphSAGE~\citep{hamilton2017inductive_sage}, and GraphSAINT~\citep{ye2024language} models, as derived from the official OGB repository\footnote{\url{https://github.com/snap-stanford/ogb}}. Additionally, the hyperparameters for GraphSAINT and the associated language models align with those specified in the GraphSAINT repository\footnote{\url{https://github.com/GraphSAINT/GraphSAINT}}. It is critical to emphasize that these hyperparameters were not individually optimized for each dataset; instead, they were uniformly applied across all three TAG datasets, in accordance with established practices. This uniform application underscores the flexibility, user-friendliness, and compatibility of our proposed method with prevailing GNN baselines.

\paragraph{Selection of Hyperparameter K.}
In our experiments, we investigated the impact of varying the hyperparameter \(K\) in Eq.\ref{eq:ICLSelec}, which controls the number of examples utilized for LLM feedback. The analysis was conducted on the \texttt{ogbn-products} dataset using 10\% of the training data. As shown in our results, setting \(K = 2\) achieved an accuracy of 79.08\%, while increasing \(K\) to 20 and 200 improved accuracy to 82.54\% and 83.09\%, respectively. However, this increase in \(K\) also led to a significant rise in computational cost, with \(K = 200\) incurring approximately 10 times the cost of \(K = 20\). Our findings indicate that \(K = 20\) offers a balanced trade-off between accuracy and computational efficiency, providing substantial performance improvements without incurring excessive computational overhead. This flexibility in adjusting \(K\) allows the method to adapt to larger datasets or resource-constrained environments. Based on this analysis, we selected \(K = 20\) as the default setting for our experiments, as it achieves an optimal balance between performance gains and computational efficiency.

\end{document}